\documentclass[conference]{IEEEtran}
\IEEEoverridecommandlockouts
\usepackage{amsmath,amssymb,amsfonts}
\usepackage{algorithmic}
\usepackage{graphicx}
\usepackage{textcomp}
\usepackage{xcolor}

\usepackage{float}
\usepackage{siunitx}
\usepackage{algorithm2e}
\usepackage{url}
\usepackage{acronym}
\usepackage{hyperref}
\hypersetup{colorlinks=true, breaklinks=true, 
  urlcolor=blue, linkcolor=blue,anchorcolor=blue,citecolor=blue,
  hypertexnames=true, final=true,
  pdfpagemode = UseNone, 
  pdfauthor = {},
  pdftitle = {},
  pdfsubject = {},
  pdfkeywords = {}
}

\graphicspath{{img/} {figs/} {./}}
\DeclareGraphicsExtensions{.pdf,.png,.jpg,.mps}

\acrodef{ADC}[ADC]{Analog-to-Digital Converter}
\acrodef{ADEXP}[AdExp-IF]{Adaptive Exponential Integrate-and-Fire}
\acrodef{ADM}[ADM]{Asynchronous Delta Modulator}
\acrodef{AE}[AE]{Address-Event}
\acrodef{AER}[AER]{Address-Event Representation}
\acrodef{AEX}[AEX]{AER EXtension board}
\acrodef{AFE}[AFE]{Analog Front-End}
\acrodef{AFM}[AFM]{Atomic Force Microscope}
\acrodef{AGC}[AGC]{Automatic Gain Control}
\acrodef{AI}[AI]{Artificial Intelligence}
\acrodef{AMDA}[AMDA]{AER Motherboard with D/A converters}
\acrodef{AMPA}[AMPA]{$\alpha$-Amino-3-hydroxy-5-methyl-4-isoxazolepropionic Acid}
\acrodef{ANN}[ANN]{Artificial Neural Network}
\acrodef{API}[API]{Application Programming Interface}
\acrodef{APMOM}[APMOM]{Alternate Polarity Metal On Metal}
\acrodef{ARM}[ARM]{Advanced RISC Machine}
\acrodef{ASIC}[ASIC]{Application Specific Integrated Circuit}
\acrodef{BCM}[BMC]{Bienenstock-Cooper-Munro}
\acrodef{BD}[BD]{Bundled Data}
\acrodef{BEOL}[BEOL]{Back-end of Line}
\acrodef{BG}[BG]{Bias Generator}
\acrodef{BMI}[BMI]{Brain-Machince Interface}
\acrodef{BTB}[BTB]{Band-to-Band tunnelling}
\acrodef{CA}[CA]{Cortical Automaton}
\acrodef{CAD}[CAD]{Computer Aided Design}
\acrodef{CAM}[CAM]{Content Addressable Memory}
\acrodef{CAVIAR}[CAVIAR]{Convolution AER Vision Architecture for Real-Time}
\acrodef{CCN}[CCN]{Cooperative and Competitive Network}
\acrodef{CDR}[CDR]{Clock-Data Recovery}
\acrodef{CFC}[CFC]{Current to Frequency Converter}
\acrodef{CHP}[CHP]{Communicating Hardware Processes}
\acrodef{CMIM}[CMIM]{Metal-Insulator-Metal Capacitor}
\acrodef{CML}[CML]{Current Mode Logic}
\acrodef{CMOL}[CMOL]{Hybrid CMOS nanoelectronic circuits}
\acrodef{CMOS}[CMOS]{Complementary Metal-Oxide-Semiconductor}
\acrodef{CNN}[CNN]{Convolutional Neural Network}
\acrodef{CNS}[CNS]{central Nervous System}
\acrodef{COTS}[COTS]{Commercial Off-The-Shelf}
\acrodef{CPG}[CPG]{Central Pattern Generator}
\acrodef{CPLD}[CPLD]{Complex Programmable Logic Device}
\acrodef{CPU}[CPU]{Central Processing Unit}
\acrodef{CSM}[CSM]{Cortical State Machine}
\acrodef{CSP}[CSP]{Constraint Satisfaction Problem}
\acrodef{CTXCTL}[CTXCTL]{CortexControl}
\acrodef{CV}[CV]{Coefficient of Variation}
\acrodef{DAC}[DAC]{Digital to Analog Converter}
\acrodef{DAS}[DAS]{Dynamic Auditory Sensor}
\acrodef{DAVIS}[DAVIS]{Dynamic and Active Pixel Vision Sensor}
\acrodef{DBN}[DBN]{Deep Belief Network}
\acrodef{DBS}[DBS]{Deep Brain Stimulation}
\acrodef{DFA}[DFA]{Deterministic Finite Automaton}
\acrodef{DIBL}[DIBL]{Drain-Induced Barrier-Lowering}
\acrodef{DI}[DI]{Delay Insensitive}
\acrodef{divmod3}[DIVMOD3]{Divisibility of a number by three}
\acrodef{DMA}[DMA]{Direct Memory Access}
\acrodef{DNF}[DNF]{Dynamic Neural Field}
\acrodef{DNN}[DNN]{Deep Neural Network}
\acrodef{DOF}[DOF]{Degrees of Freedom}
\acrodef{DPE}[DPE]{Dynamic Parameter Estimation}
\acrodef{DPI}[DPI]{Differential Pair Integrator}
\acrodef{DRAM}[DRAM]{Dynamic Random Access Memory}
\acrodef{DR}[DR]{Dual Rail}
\acrodef{DRRZ}[DR-RZ]{Dual-Rail Return-to-Zero}
\acrodef{DSP}[DSP]{Digital Signal Processor}
\acrodef{DVS}[DVS]{Dynamic Vision Sensor}
\acrodef{DYNAP}[DYNAP]{Dynamic Neuromorphic Asynchronous Processor}
\acrodef{EBL}[EBL]{Electron Beam Lithography}
\acrodef{ECG}[ECG]{Electrocardiography}
\acrodef{ECoG}[ECoG]{Electrocorticography}
\acrodef{EDVAC}[EDVAC]{Electronic Discrete Variable Automatic Computer}
\acrodef{EEG}[EEG]{Electroencephalography}
\acrodef{EIN}[EIN]{Excitatory-Inhibitory Network}
\acrodef{EM}[EM]{Expectation Maximization}
\acrodef{EMG}[EMG]{Electromyography}
\acrodef{EOG}[EOG]{Electrooculogram}
\acrodef{EPSC}[EPSC]{Excitatory Post-Synaptic Current}
\acrodef{EPSP}[EPSP]{Excitatory Post-Synaptic Potential}
\acrodef{EZ}[EZ]{Epileptogenic Zone}
\acrodef{FDSOI}[FDSOI]{Fully-Depleted Silicon on Insulator}
\acrodef{FET}[FET]{Field-Effect Transistor}
\acrodef{FFT}[FFT]{Fast Fourier Transform}
\acrodef{FI}[F-I]{Frequency--Current}
\acrodef{FMA}[FMA]{Floating Microelectrode Array} 
\acrodef{FNN}[FNN]{Feed-forward Neural Network}
\acrodef{FPGA}[FPGA]{Field Programmable Gate Array}
\acrodef{FR}[FR]{Fast Ripple}
\acrodef{FSA}[FSA]{Finite State Automaton}
\acrodef{FSM}[FSM]{Finite State Machine}
\acrodef{GABA}[GABA]{$\gamma$-Aminobutanoic Acid}
\acrodef{GIDL}[GIDL]{Gate-Induced Drain Leakage}
\acrodef{GOPS}[GOPS]{Giga-Operations per Second}
\acrodef{GPIO}[GPIO]{General Purpose I/O}
\acrodef{GPU}[GPU]{Graphical Processing Unit}
\acrodef{GT}[GT]{Ground Truth}
\acrodef{GUI}[GUI]{Graphical User Interface}
\acrodef{HAL}[HAL]{Hardware Abstraction Layer}
\acrodef{HFO}[HFO]{High Frequency Oscillation}
\acrodef{HH}[H\&H]{Hodgkin \& Huxley}
\acrodef{HMM}[HMM]{Hidden Markov Model}
\acrodef{HR}[HR]{Human Readable}
\acrodef{HRS}[HRS]{High-Resistive State}
\acrodef{HSE}[HSE]{Handshaking Expansion}
\acrodef{HW}[HW]{Hardware}
\acrodef{hWTA}[hWTA]{Hard Winner-Take-All}
\acrodef{IC}[IC]{Integrated Circuit}
\acrodef{ICT}[ICT]{Information and Communication Technology}
\acrodef{iEEG}[iEEG]{Intracranial Electroencephalography}
\acrodef{IF2DWTA}[IF2DWTA]{Integrate \& Fire 2-Dimensional WTA}
\acrodef{IF}[I\&F]{Integrate-and-Fire}
\acrodef{IFSLWTA}[IFSLWTA]{Integrate \& Fire Stop Learning WTA}
\acrodef{IMU}[IMU]{Inertial Measurement Unit}
\acrodef{INCF}[INCF]{International Neuroinformatics Coordinating Facility}
\acrodef{INI}[INI]{Institute of Neuroinformatics}
\acrodef{IO}[I/O]{Input/Output}
\acrodef{IoT}[IoT]{Internet of Things}
\acrodef{IP}[IP]{Intellectual Property}
\acrodef{IPSC}[IPSC]{Inhibitory Post-Synaptic Current}
\acrodef{IPSP}[IPSP]{Inhibitory Post-Synaptic Potential}
\acrodef{ISI}[ISI]{Inter-Spike Interval}
\acrodef{JFLAP}[JFLAP]{Java - Formal Languages and Automata Package}
\acrodef{LEDR}[LEDR]{Level-Encoded Dual-Rail}
\acrodef{LFP}[LFP]{Local Field Potential}
\acrodef{LIFE}[LIFE]{Longitudinal Intrafascicular Electrodes}
\acrodef{LIF}[LI\&F]{Leaky Integrate-and-Fire}
\acrodef{LLC}[LLC]{Low Leakage Cell}
\acrodef{LNA}[LNA]{Low-Noise Amplifier}
\acrodef{LPF}[LPF]{Low Pass Filter}
\acrodef{LRS}[LRS]{Low-Resistive State}
\acrodef{LSM}[LSM]{Liquid State Machine}
\acrodef{LTD}[LTD]{Long Term Depression}
\acrodef{LTI}[LTI]{Linear Time-Invariant}
\acrodef{LTP}[LTP]{Long Term Potentiation}
\acrodef{LTU}[LTU]{Linear Threshold Unit}
\acrodef{LUT}[LUT]{Look-Up Table}
\acrodef{LVDS}[LVDS]{Low Voltage Differential Signaling}
\acrodef{MCMC}[MCMC]{Markov-Chain Monte Carlo}
\acrodef{MEA}[MEA]{Multielectrode Arrays}
\acrodef{MEMS}[MEMS]{Micro Electro Mechanical System}
\acrodef{MFR}[MFR]{Mean Firing Rate}
\acrodef{MIM}[MIM]{Metal Insulator Metal}
\acrodef{ML}[ML]{Machine Learning}
\acrodef{MLP}[MLP]{Multilayer Perceptron}
\acrodef{MOSCAP}[MOSCAP]{Metal Oxide Semiconductor Capacitor}
\acrodef{MOSFET}[MOSFET]{Metal Oxide Semiconductor Field-Effect Transistor}
\acrodef{MOS}[MOS]{Metal Oxide Semiconductor}
\acrodef{MRI}[MRI]{Magnetic Resonance Imaging}
\acrodef{NCS}[NCS]{Neuromorphic Cognitive Systems}
\acrodef{NDFSM}[NDFSM]{Non-deterministic Finite State Machine} 
\acrodef{ND}[ND]{Noise-Driven}
\acrodef{NEF}[NEF]{Neural Engineering Framework}
\acrodef{NHML}[NHML]{Neuromorphic Hardware Mark-up Language}
\acrodef{NIL}[NIL]{Nano-Imprint Lithography}
\acrodef{NI}[NI]{Neural Interface}
\acrodef{NMDA}[NMDA]{\textit{N}-Methyl-\textsc{d}-aspartate}
\acrodef{NME}[NE]{Neuromorphic Engineering}
\acrodef{NN}[NN]{Neural Network}
\acrodef{NOC}[NoC]{Network-on-Chip}
\acrodef{NRZ}[NRZ]{Non-Return-to-Zero}
\acrodef{NSM}[NSM]{Neural State Machine}
\acrodef{OR}[OR]{Operating Room}
\acrodef{OTA}[OTA]{Operational Transconductance Amplifier}
\acrodef{PCB}[PCB]{Printed Circuit Board}
\acrodef{PCHB}[PCHB]{Pre-Charge Half-Buffer}
\acrodef{PCM}[PCM]{Phase Change Memory}
\acrodef{PC}[PC]{Personal Computer}
\acrodef{PDK}[PDK]{Process Design Kit}
\acrodef{PE}[PE]{Phase Encoding}
\acrodef{PFA}[PFA]{Probabilistic Finite Automaton}
\acrodef{PFC}[PFC]{Prefrontal Cortex}
\acrodef{PFM}[PFM]{Pulse Frequency Modulation}
\acrodef{PNI}[PNI]{Peripheral Nerve Interface}
\acrodef{PNS}[PNS]{Peripheral Nervous System}
\acrodef{PPG}[PPG]{Photoplethysmography}
\acrodef{PR}[PR]{Production Rule}
\acrodef{PSC}[PSC]{Post-Synaptic Current}
\acrodef{PSP}[PSP]{Post-Synaptic Potential}
\acrodef{PSTH}[PSTH]{Peri-Stimulus Time Histogram}
\acrodef{PV}[PV]{Parvalbumin}
\acrodef{QDI}[QDI]{Quasi Delay Insensitive}
\acrodef{RAM}[RAM]{Random Access Memory}
\acrodef{RA}[RA]{Resected Area}
\acrodef{RDF}[RDF]{Random Dopant Fluctuation}
\acrodef{RELU}[ReLu]{Rectified Linear Unit}
\acrodef{RLS}[RLS]{Recursive Least-Squares}
\acrodef{RMSE}[RMSE]{Root Mean Square-Error}
\acrodef{RMS}[RMS]{Root Mean Square}
\acrodef{RNN}[RNN]{Recurrent Neural Network}
\acrodef{ROLLS}[ROLLS]{Reconfigurable On-Line Learning Spiking}
\acrodef{RRAM}[R-RAM]{Resistive Random Access Memory}
\acrodef{R}[R]{Ripple}
\acrodef{RISC}[RISC]{Reduced Instruction Set Computer}
\acrodef{RSA}[RSA]{Respiratory Sinus Arrhythmia}
\acrodef{SAC}[SAC]{Selective Attention Chip}
\acrodef{SAT}[SAT]{Boolean Satisfiability Problem}
\acrodef{SCI}[SCI]{Spinal Cord Injury}
\acrodef{SCX}[SCX]{Silicon CorteX}
\acrodef{SD}[SD]{Signal-Driven}
\acrodef{SEM}[SEM]{Spike-based Expectation Maximization}
\acrodef{SLAM}[SLAM]{Simultaneous Localization and Mapping}
\acrodef{SNN}[SNN]{Spiking Neural Network}
\acrodef{SNR}[SNR]{Signal to Noise Ratio}
\acrodef{SOC}[SoC]{System-On-Chip}
\acrodef{SOI}[SOI]{Silicon on Insulator}
\acrodef{SOZ}[SOZ]{Seizure Onset Zone}
\acrodef{SP}[SP]{Separation Property}
\acrodef{SPI}[SPI]{Serial Peripheral Interface}
\acrodef{SRAM}[SRAM]{Static Random Access Memory}
\acrodef{SST}[SST]{Somatostatin}
\acrodef{STDP}[STDP]{Spike-Timing Dependent Plasticity}
\acrodef{STD}[STD]{Short-Term Depression}
\acrodef{STP}[STP]{Short-Term Plasticity}
\acrodef{STT-MRAM}[STT-MRAM]{Spin-Transfer Torque Magnetic Random Access Memory}
\acrodef{STT}[STT]{Spin-Transfer Torque}
\acrodef{SVM}[SVM]{Support Vector Machine}
\acrodef{SW}[SW]{Software}
\acrodef{sWTA}[sWTA]{soft Winner-Take-All}
\acrodef{TCAM}[TCAM]{Ternary Content-Addressable Memory}
\acrodef{TFT}[TFT]{Thin Film Transistor}
\acrodef{TIME}[TIME]{Transverse Intrafascicular Multichannel Electrode}
\acrodef{TLE}[TLE]{Temporal Lobe Epilepsy}
\acrodef{UEA}[UEA]{Utah Electrode Array}
\acrodef{USB}[USB]{Universal Serial Bus}
\acrodef{USEA}[USEA]{Utah Slanted Electrode Array}
\acrodef{VHDL}[VHDL]{VHSIC Hardware Description Language}
\acrodef{VHSIC}[VHSIC]{Very High Speed Integrated Circuits}
\acrodef{VIP}[VIP]{Vasoactive Intestinal Peptide}
\acrodef{VLSI}[VLSI]{Very Large Scale Integration}
\acrodef{VNS}[VNS]{Vagal Nerve Stimulation}
\acrodef{VOR}[VOR]{Vestibulo-Ocular Reflex}
\acrodef{VSA}[VSA]{Vector Symbolic Architecture}
\acrodef{WCST}[WCST]{Wisconsin Card Sorting Test}
\acrodef{WTA}[WTA]{Winner-Take-All}
\acrodef{XML}[XML]{eXtensible Mark-up Language}

\begin{document}

\newcommand{\refs}{\textcolor{red}{[\dots]}}
\newcommand{\rdots}{\textcolor{red}{\dots}}
\newcommand{\bs}[1]{\boldsymbol{#1}}

\title{Genetic Motifs as a Blueprint for Mismatch-Tolerant Neuromorphic Computing\\
}

\author{\IEEEauthorblockN{Tommaso Boccato$^1$, Dmitrii Zendrikov$^3$, Nicola Toschi$^{1, 2}$, Giacomo Indiveri$^3$}
\IEEEauthorblockA{$^1$\textit{Department of Biomedicine and Prevention, University of Rome Tor Vergata, Rome, Italy}\\
$^2$\textit{A.A. Martinos Center for Biomedical Imaging and Harvard Medical School, Boston, USA}\\
$^3$\textit{Institute of Neuroinformatics, University of Zurich and ETH Zurich, Zurich, Switzerland}\\
email: \texttt{tommaso.boccato@uniroma2.it}}
}

\maketitle

\begin{abstract}
  Mixed-signal implementations of \acp{SNN} offer a promising solution to edge computing applications that require low-power and compact embedded processing systems.
  However, device mismatch in the analog circuits of these neuromorphic processors poses a significant challenge to the deployment of robust processing in these systems.
  Here we introduce a novel architectural solution inspired by biological development to address this issue.
  Specifically we propose to implement architectures that incorporate network motifs found in developed brains through a differentiable re-parameterization of weight matrices based on gene expression patterns and genetic rules.
  Thanks to the gradient descent optimization compatibility of the method proposed, we can apply the robustness of biological neural development to neuromorphic computing.

  To validate this approach we benchmark it using the Yin-Yang classification dataset, and compare its performance with that of standard multilayer perceptrons trained with state-of-the-art hardware-aware training method.
  Our results demonstrate that the proposed method mitigates mismatch-induced noise without requiring precise device mismatch measurements, effectively outperforming alternative hardware-aware techniques proposed in the literature, and providing a more general solution for improving the robustness of \acp{SNN} in neuromorphic hardware.
\end{abstract}

\begin{IEEEkeywords}
spiking neural networks, mixed-signal chips, device mismatch, network neuroscience
\end{IEEEkeywords}

\acresetall

\section{Introduction}\label{sec:introduction}

In today's \ac{AI} landscape, state-of-the-art models are becoming increasingly energy-intensive~\cite{DBLP:journals/corr/abs-1906-02243}, and 
demand for energy-efficient solutions is rising.
Within this context, mixed-signal implementations of \acp{SNN} that use analog circuits and the physics of silicon to directly emulate the computational properties of biological neural networks represent a promising low-power technology~\cite{Indiveri_Sandamirskaya19}.
Several neuromorphic processors of this type have been recently developed~\cite{Richter_etal24,Moradi_etal18,10.3389/fnins.2022.795876,Neckar_etal19,Giulioni_etal08,Chicca_etal03}.

These processors, characterized by analog circuits that implement the neural dynamics and digital ones for the routing and interfacing logic, can achieve ultra-low-power and ultra-low-latency capabilities, by operating analog transistors in the sub-threshold regime and digital ones in an asynchronous, self-clocked mode~\cite{Liu_etal02a,Liu_etal14}.
However, mainly due to the device-mismatch problems in their analog circuits, neural networks built using these processors are often less accurate than their \ac{AI} \ac{ANN} counterparts, implemented on standard digital computing technologies. 

To successfully deploy these types of devices in real-world applications it is necessary to find a way to achieve negligible mismatch-induced performance loss, tighter control on the dynamics of the models, and possibly ease in training models for tasks of interest.
Several solutions have been proposed to mitigate device mismatch in neuromorphic systems~\cite{Cameron_Murray08,Neftci_Indiveri10,Bamford_etal13,Buchel_etal21a,Cakal_2024}, however most rely on calibration techniques that require to measure the device mismatch effects in every single device, require precise characterization of the noise in the circuits, or can be applied only to a fixed set of neural network architectures (e.g., only recurrent networks).

It seems natural, then, to shift toward a more general architectural solution.
A first hint at the potential of employing ad-hoc network topologies is provided in~\cite{NEURIPS2019_e9874147}, where weight-agnostic neural networks are proposed.
In this family of models, all synaptic weights share the exact same value, which can be sampled from a distribution defined a-priori, and the performance of the networks remains constant regardless of the value.
Unfortunately, the approach is effective only when individual activation functions can be selected from a diverse set of functions (e.g., linear, sigmoid, Gaussian, sinusoid); this is not the case for neuromorphic chips where the activation functions are directly inherited from the neuron model implemented in silicon.

Following the neuromorphic engineering principles, the novel solution we propose, draws inspiration from nature.
Indeed, most biological processes are also very often ``noisy''.
A notable example is the stochasticity that affects synaptogenesis--the process of synapse formation--during brain development.
There exist probabilities according to which neurons can connect with compatible post-synaptic partners.
Despite this, it is possible to identify ``meta-topologies'' or ``probabilistic skeletons''~\cite{boccato2024optimizing,Stockl2021.05.18.444689} from which perfectly functional \acp{NN} can be sampled, or in biological terms, developed.
Inspired by this, our proposal involves developing a spiking neural architecture\footnote{The code will be available soon at \url{https://github.com/BoCtrl-C}.}, compatible with gradient descent-based optimizers, characterized by the same network motifs evolution could have optimized over millions of years to let developed brains be robust to synaptogenesis stochasticity and other forms of biological noise.

\section{Methods}

\begin{figure}
\centerline{\includegraphics[width=.8\columnwidth]{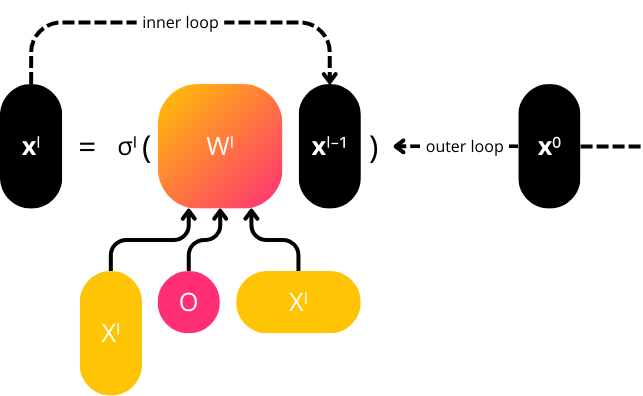}}
\caption{Our spiking neural architecture, which exhibits mismatch-tolerant computing capabilities through bio-inspired network motifs introduced via re-parameterization of the weight matrices.}
\label{fig:architecture}
\end{figure}

\begin{figure}
\centerline{\includegraphics[width=.7\columnwidth]{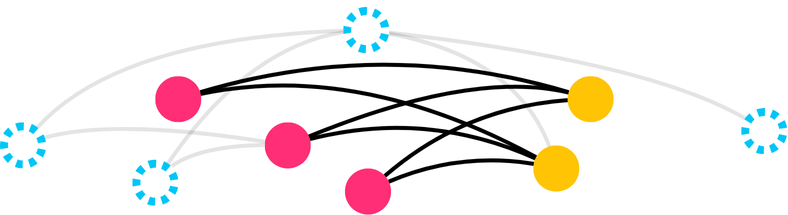}}
\caption{A fully-connected bi-partite motif example, where two clusters of neurons--the \textcolor[HTML]{ff3076}{$\bullet$} and \textcolor[HTML]{ffc300}{$\bullet$} nodes--have no intra-cluster connections but are fully connected between clusters.}
\label{fig:motif}
\end{figure}

The idea that in animals the neural circuits responsible for functions which are critical for survival are mostly decoded from the genome~\cite{zador_critique_2019} is spreading rapidly among neuroscientists.
In this context, the Connectome Model (CM)~\cite{doi:10.1073/pnas.2009093117} was developed as a computational tool for unveiling the genetic mechanisms that govern neuronal wiring.
According to the CM, synapses result from the interaction of proteins translated from genes that are selectively expressed in individual neurons.
Formally, a binary connectome $B \in \{0, 1\}^{N \times N}$ can be obtained from:
\begin{equation}\label{eq:cm}
    B = \mathcal{H}\big(XOX^T\big)
\end{equation}
where $X \in \{0, 1\}^{N \times G}$ and $O \in [0, 1]^{G \times G}$ are matrix representations of gene expression patterns and genetic rules (i.e., interaction probabilities between proteins), $\mathcal{H}$ denotes the Heaviside function, while $N$ and $G$ represent the number of neurons and genes, respectively.
Specifically, $X$ is a binary matrix in which the generic entry $X_{ui}$ tells us if gene $i$ is expressed or not in neuron $u$ while $O_{ij}$ represents the probability of forming synapses when genes $i$ and $j$ are involved in the process.
Interestingly, the CM induces a very specific topology in the resulting connectome; this can be seen by making the matrix multiplications in \eqref{eq:cm} explicit:
\begin{align}\label{eq:cm-entry}
    B_{uv} &= \mathcal{H}\bigg(\sum_i X_{ui} \sum_j O_{ij}X_{jv}^T\bigg)\nonumber\\
    &= \mathcal{H}\bigg(\sum_{i, j} X_{ui}O_{ij}X_{vj}\bigg)
\end{align}
According to \eqref{eq:cm-entry}, indeed, every time there exist two neuron clusters $\{u : X_{uk} = 1\} \neq \emptyset$ and $\{v : X_{vl} = 1\} \neq \emptyset$ whose corresponding genetic rule $O_{kl} > 0$, a fully-connected bi-partite motif (see Figure \ref{fig:motif}) can be identified in the connectome graph: $\sum_{i, j} X_{ui}O_{ij}X_{vj} \ge O_{kl} > 0 \Rightarrow B_{uv} = 1$.
Hence, the overall graph results from the union of all genetically-induced motifs.

We therefore considered the CM as a starting point for including the aforementioned motifs into the proposed spiking neural architecture, pursuing the objective of transferring the robustness exhibited by biological neuronal networks to the world of neuromorphic computing.
In order to do so, we relaxed the $X$ and $O$ matrices to $\mathbb{R}$~\cite{barabasi_complex_2023} and replaced each weight matrix of a feedforward layered network of leaky integrate-and-fire (LIF) neurons with the decomposition in \eqref{eq:cm}, Heaviside function excluded.

Specifically, our \acp{LIF} operate according to the following temporal dynamics:
\begin{equation}\label{eq:neuron-model}
    U(t) = \begin{cases}
        \beta U(t - 1) + RI_{in}(t),\ &\text{if}\ U(t - 1) \le U_{thr}\\
        RI_{in}(t) &\text{otherwise}
    \end{cases}
\end{equation}
where $U(t)$ is the membrane potential of the neuron at time $t$, $\beta$ is a decay factor, $I_{in}$ represents the current entering the neuron, $R$ is a ``virtual'' resistance\footnote{For convenience, $R$ is set to \SI{1}{\ohm}/\SI{}{\milli\ohm}/\dots regardless of the scale according to which the other quantities are expressed.} to make the current term a voltage and $U_{thr}$ denotes the neuron threshold.
The relationship between $\beta$ and the neuron time constant $\tau$ is $\beta = 1 - \frac{\Delta t}{\tau}$, with $\Delta t$ defined as the simulation time constant.
The neuron emits a spike when the $U_{thr}$ threshold is exceeded:
\begin{equation}\label{eq:spiking-mechanism}
    S(t) = \begin{cases}
        1,\ &\text{if}\ U(t) > U_{thr}\\
        0 &\text{otherwise}
    \end{cases}
\end{equation}
Proceeding with the description of our architecture, but at the network level, the neuron input current can be further specified as the summation of all spikes coming from the pre-synaptic neurons modulated through the respective synaptic conductances, $\bs{w}$: $I_{in}(t) = \bs{w}^T\bs{x}(t)$.
Here, entry $\bs{w}_u$ stores the conductance value of the connection between pre-synaptic neuron $u$ and the post-synaptic neuron, while $\bs{x}_u \in \{0, 1\}$ is the output of neuron $u$\footnote{There is no relationship between the $X$ notation used for gene expression patterns and the one used for neuron outputs, $\bs{x}$.}.
In a multi-partite topology made of fully-connected graphs--a multilayer perceptron (MLP)--like the one who characterizes our architecture, it is possible to compute all inputs for the neuronal layer $l$ in one single operation: $W^l\bs{x}(t)$; this is done by stacking vertically the row vectors $\bs{w}^T$ for all neurons $v$ in layer $l$.
It is precisely the weight matrix $W^l$ that can be re-parameterized exploiting \eqref{eq:cm}:
\begin{equation}\label{eq:reparameterization}
    W^l = X^lO{X^{l - 1}}^T
\end{equation}
where $X^l$ collects the rows of $X$ corresponding to neurons in $l$.
Assuming that $\Delta t$ is much larger than the time required to propagate signals throughout the \acp{NN} built with our architecture, their forward pass can be implemented following the pseudocode in Algorithm \ref{alg:forward-pass}.
\begin{algorithm}
\caption{Pseudocode for our forward pass.}\label{alg:forward-pass}
\small
$U_u(0) = 0,\ \forall u$\;
\For{$t \gets 1$ \upshape{to} $T$}{
    \For{$l \gets 1$ \upshape{to} $H - 1$}{
        $\bs{x}^l(t) \gets \sigma^l\Big(\big(X^lO{X^{l - 1}}^T\big)\bs{x}^{l - 1}(t)\Big)$\;
    }
}
\Return $[\bs{x}^{H - 1}(1), \dots, \bs{x}^{H - 1}(T)]$\;
\vspace{.25em}
\end{algorithm}
In the algorithm, $U_u(0)$ is the membrane potential of neuron $u$ at time 0, $T$ represents the simulation duration, $H$ is the number of layers in the network, $\bs{x}^0(t)$ denotes the input signals (one for each input neuron), and $\sigma^l(\bullet) = [\dots, S_u(t), \dots]^T$.
In this last equation, $\sigma$ takes as implicit argument also the time step at which it is applied and $S_u(t)$ represents the spiking mechanism of neuron $u$, with $u$ belonging to layer $l$.
Also $S_u$ requires an implicit update of $U_u$ and depends on the neuron input.
Summarizing, the forward pass resets the membrane potentials and, for each simulation time step, processes the respective input spikes while updating the neuron states.

Thanks to the exploitation of surrogate gradients~\cite{Neftci_etal19} for the functions in \eqref{eq:neuron-model} and \eqref{eq:spiking-mechanism}, and to the re-parameterization in \eqref{eq:reparameterization}, our architecture never breaks differentiability w.r.t.
the entries of $X$ and $O$; this makes it possible to treat gene expression patterns and genetic rules as learnable parameters that can be optimized with gradient-based techniques.
An overview of the architecture is shown in Figure \ref{fig:architecture}.

\section{Experiments \& Results}

In this section, we describe the experiments conducted to validate the proposed architecture.
We begin with a description of the chosen benchmark task, continue with the experimental protocol, and conclude by presenting the obtained results.

\subsection{Dataset}

\begin{figure}
\centerline{\includegraphics[width=.9\columnwidth]{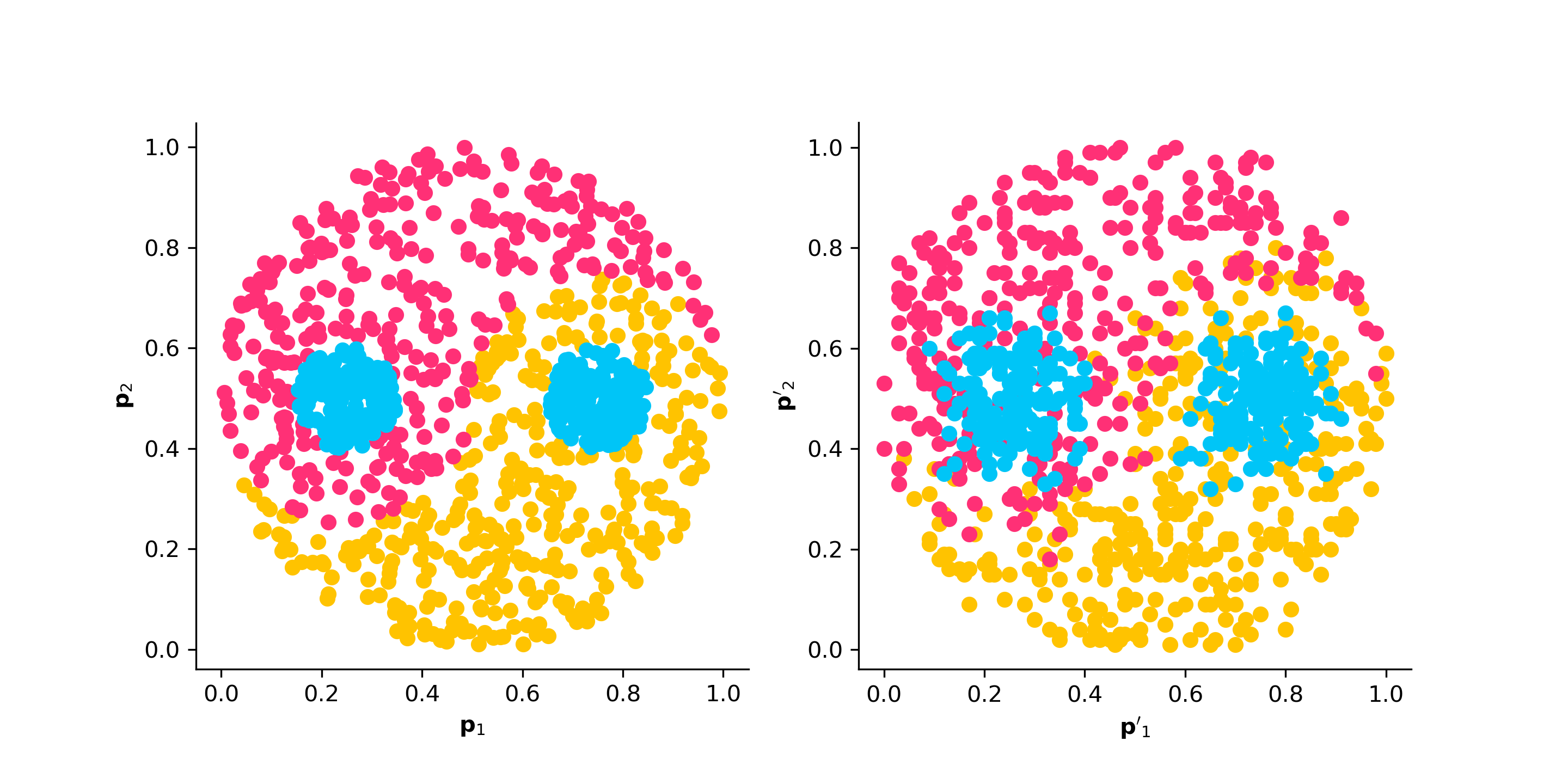}}
\caption{The Yin-Yang classification dataset~\cite{10.1145/3517343.3517380}.
\textbf{Left}: the original dataset.
\textbf{Right}: a visual representation of our rate encoded version.}
\label{fig:dataset}
\end{figure}

\begin{figure*}
\centerline{\includegraphics[width=.7\textwidth]{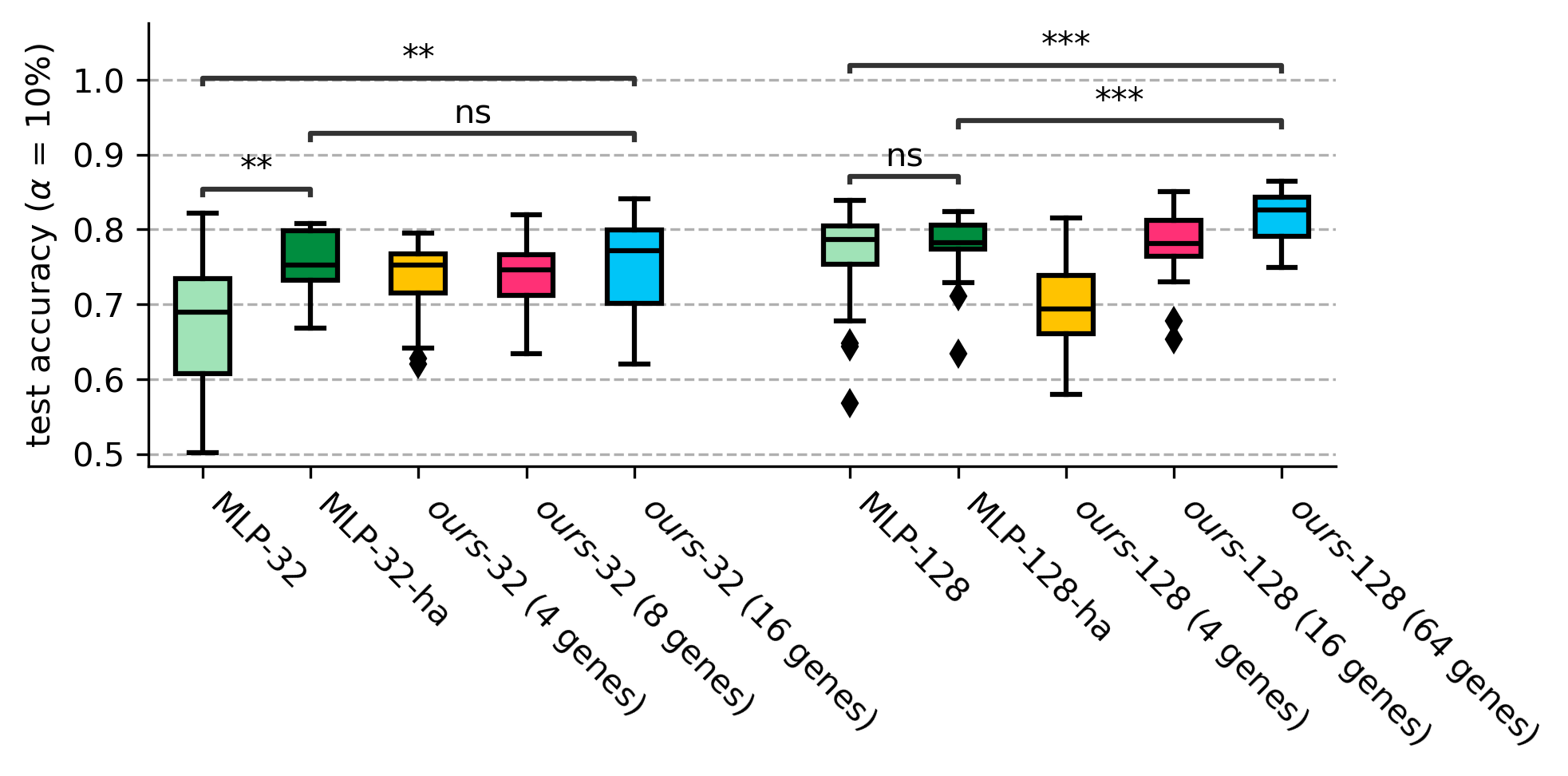}}
\caption{A box plot showing the accuracy distributions of models tested in our simulated on-chip deployment.
Noise injection is characterized by a coefficient of variation of 10\% ($\alpha = 0.1$).
The two groups of boxes represent different network sizes, and models are color-coded.
Statistical annotations are provided at the top of the plot.}
\label{fig:results}
\end{figure*}

We chose the Yin-Yang dataset~\cite{10.1145/3517343.3517380} for classification as our reference benchmark for evaluating the performance of the proposed architecture.
The dataset consists of three clusters of 2D points, one for each classification label, that together depict the Yin-Yang symbol (see the left panel of Figure \ref{fig:dataset}).
Samples are provided within three different splits--``training'', ``validation'' and ``test''--of size 5000, 1000 and 1000.
Unlike related benchmarks, data have been constructed in such a way as to be simple enough for speeding-up model trainings, but at the same time sufficiently complex to let performance differences emerge when testing different kinds of classifiers.
Moreover, the dataset samples present two additional features--four in total--generated by a mirrored axis.
The obtained symmetry is indeed helpful when training networks without neuron biases.

In order to make dataset samples compatible with the developed spiking architecture, we rely on a standard rate encoding: $\bs{x}_i^0(t) \sim \mathcal{B}(\bs{p}_i),\ \forall t$.
In this last equation, $\bs{x}_i^0(t)$ represents the $i$-th network input at time $t$, $\mathcal{B}$ is a Bernoulli random variable and $\bs{p}_i$ denotes the $i$-th coordinate--or feature--of the considered dataset point.
It is worth noting that the encoding process is responsible for injecting in the samples two different types of noise.
The former is linked to the inherent stochasticity of rate encoding where the same coordinates can be transformed in different input signals.
The latter, instead, can be considered as a quantization noise and depends on the simulation duration $T$.
We make this clear in the right panel of Figure \ref{fig:dataset} where each example point is drawn in accordance with the coordinates expressed by: $\bs{p}'_i = \frac{1}{T}\sum_t \bs{x}_i^0(t),\ i = 1, 2$.
As a consequence, $T$ fixes the number of values that a coordinate can assume.

\subsection{Experiments}
In order to assess the competitiveness of our architecture, we compared the performance of our models with two baselines on a simulated on-chip deployment.

The first baseline is represented by a standard \ac{MLP} in which the weight matrices are learned directly, with no re-parameterizations.
The second baseline, instead, corresponds to a modified \ac{MLP} whose forward pass has been updated to be compatible with the hardware-aware training procedure developed by Cakal et al.
\cite{Cakal_2024} and discussed in Section \ref{sec:introduction}.
This change is detailed in Algorithm \ref{alg:forward-pass-cakal}.
\begin{algorithm}
\caption{Pseudocode for the forward pass in~\cite{Cakal_2024}.}\label{alg:forward-pass-cakal}
\small
$E^l \sim \mathcal{N}\big(W^l, \alpha^2(W^l \odot W^l)\big),\ \forall l$\;
$U_u(0) = 0,\ \forall u$\;
\For{$t \gets 1$ \upshape{to} $T$}{
    \For{$l \gets 1$ \upshape{to} $H - 1$}{
        $\bs{x}^l(t) \gets \sigma^l\Big(\big(W^l + E^l\big)\bs{x}^{l - 1}(t)\Big)$\;
    }
}
\Return $[\bs{x}^{H - 1}(1), \dots, \bs{x}^{H - 1}(T)]$\;
\vspace{.25em}
\end{algorithm}
At the beginning of the forward pass, the mismatch-induced noise is simulated by sampling, for each synaptic weight, an ``error'' value from a Gaussian random variable with a mean equal to the weight itself and a standard deviation equal to the absolute value of the weight, multiplied by a predetermined coefficient of variation, $\alpha$.
In the pseudocode, the $\mathcal{N}$ notation for the independent Gaussian random variables must be intended as point-wise, and $E^l$ has the same shape of $W^l$.
Then, the produced noise matrices are added, inside the inner loop, to the respective weight matrices every time the synaptic weights are used in computations.
Also in the context of this baseline, the $W^l$ matrices undergo no further decomposition and synaptic weights are learned directly.

We conducted two experiments, each with its own set of \acp{SNN} characterized by a single hidden layer of 32 or 128 neurons.
In both experiments, we trained with Adam~\cite{DBLP:journals/corr/KingmaB14} ($\beta_1 = 0.9$, $\beta_2 = 0.999$), and for 300 epochs (512 samples per batch), three models instantiated from our architecture ($G = 4, 8, 16$ or $G = 4, 16, 64$ depending on the experiment) and two spiking \acp{MLP}--one for each baseline--on the Yin-Yang dataset's training split.
A Cross Entropy Spike Count Loss~\cite{Neftci_etal19} was minimized.
Several learning rates (0.03, 0.003, 0.0003) were explored and we selected for the testing phase the models which achieved the lowest loss on the validation split.
For all forward passes, we set $\Delta t = 1$, $T = 100$ and $\beta = 0.9 \Rightarrow \tau = 10$.

Our testing phase simulates faithfully the on-chip deployment of the models, by evaluating their performance in the presence of mismatch-induced noise.
We did so by enabling the noise injection of Algorithm \ref{alg:forward-pass-cakal} in all networks' forward passes.
For each trained network, we predicted the dataset's test split labels 30 times, independently.
Also the error matrices (i.e., $\{E^l\}$) were sampled in total 30 times, therefore simulating 30 different chips.
The coefficient of variation $\alpha$ was set to 0.1 (10\%), a plausible value for mixed-signal neuromorphic processors~\cite{Zendrikov_2023}.

\subsection{Results}

We show our results in Figure \ref{fig:results}, where the group of boxes on the left refers to the networks with 32 neurons in the hidden layer while the group on the right to the ones with 128 neurons.
We underline again how all models in the same group share the same number of synaptic weights.

Some important trends emerge from the data.
First of all, our best models--in terms of median test accuracy--either outperform (in a statistical sense) or perform on par with the \acp{MLP} trained according to the hardware-aware procedure.
It is important to highlight that, in both experiments, the best networks are the ones characterized by the highest number of genes.
In general, the higher the number of genes the better the performance achieved by the architecture proposed.
The 4 genes-8 genes pair inside the first group of models represents an exception to this rule of thumb, but the medians of the respective accuracy distributions do not differ significantly in a statistical sense.
Also the models of Cakal et al., in turn, outperform or perform on par with the \ac{MLP} baselines; our networks, however, always surpass the accuracy achieved by these baselines.
Overall, it seems beneficial to rely on networks characterized by a larger hidden layer; on the other hand, this also means optimizing a higher number of parameters.
We finally specify that all comparisons were carried out through Mann-Whitney statistical tests with Bonferroni correction.

\section{Conclusions}

We proposed a new approach that addresses device mismatch in mixed-signal neuromorphic chips, by synthesizing spiking neural architectures which incorporate network motifs found in developed biological brains.
We integrate these network motifs into the \acp{SNN} architecture by re-parameterizing their weight matrices using a function that 
is differentiable w.r.t.
learnable parameters, which represent gene expression patterns and genetic rules. 

Our experiments on the chosen benchmark demonstrate that our models effectively mitigate the effects of mismatch-induced noise, outperforming the current state-of-the-art in classification accuracy.
Despite requiring two additional matrix multiplications per layer compared to standard \acp{MLP} during the off-line training forward pass, a key advantage of our method is that it does not require precise characterization of device mismatch, unlike existing hardware-aware training techniques.

Future work will focus on exploring the impact of weight quantization in these networks and on validating the results found in these studies through deployment on mixed-signal neuromorphic hardware.



\bibliographystyle{ieeetr}
\bibliography{refs,biblioncs}

\end{document}